\documentclass[letterpaper, 10 pt, conference]{ieeeconf}  % Comment this line out if you need a4paper

\usepackage{dcolumn}
\usepackage{graphicx}
\let\labelindent\relax
\usepackage{enumitem}

\newlist{condenum}{enumerate}{1} % 'condenum': a new, enumerate-like list env.
\setlist[condenum]{label=\bfseries C\arabic*.,
	ref=\arabic*, wide}

\usepackage{siunitx,booktabs}
\usepackage{subcaption}
\usepackage{multicol} 
\usepackage{amsmath}
\usepackage{amssymb}
\usepackage{balance}

\graphicspath{{figures/}}

\begin{document}

\title{\LARGE \bf
 Anticipating Human Intention for Full-Body Motion Prediction in Object Grasping and Placing Tasks
}

\author{Philipp Kratzer$^{1,2}$, Niteesh Balachandra Midlagajni$^2$, Marc Toussaint$^{3}$ and Jim Mainprice$^{1,2}$\\% <-this % stops a space
\authorblockA{$^1$\tt{\small{firstname.lastname@ipvs.uni-stuttgart.de}}}
\authorblockA{$^1$Machine Learning and Robotics Lab, University of Stuttgart, Germany}
\authorblockA{$^2$Humans to Robots Motions Research Group ; HRM ; University of Stuttgart, Germany}
\authorblockA{$^3$Learning and Intelligent Systems Lab ;  Technical University of Berlin, Germany}
\vspace{-1.2cm}
}

\twocolumn[{%
\renewcommand\twocolumn[1][]{#1}%
\maketitle
\begin{center}
  \centering
  \newcommand{\ltscale}{.30}
\centering
\includegraphics[width=.32\linewidth]{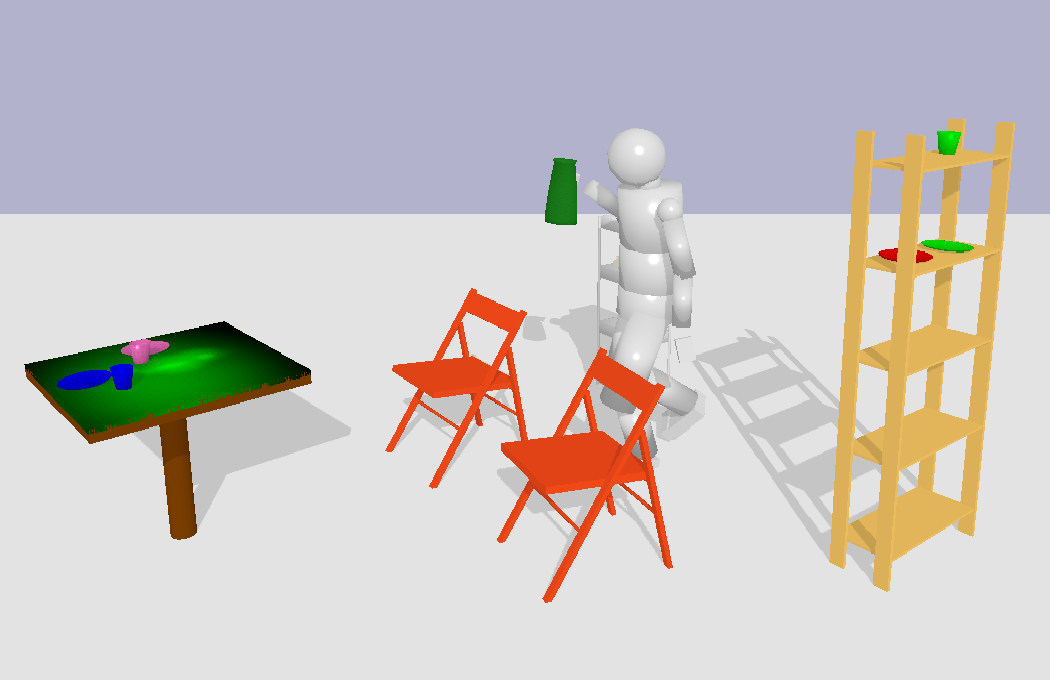}
\centering
\includegraphics[width=.32\linewidth]{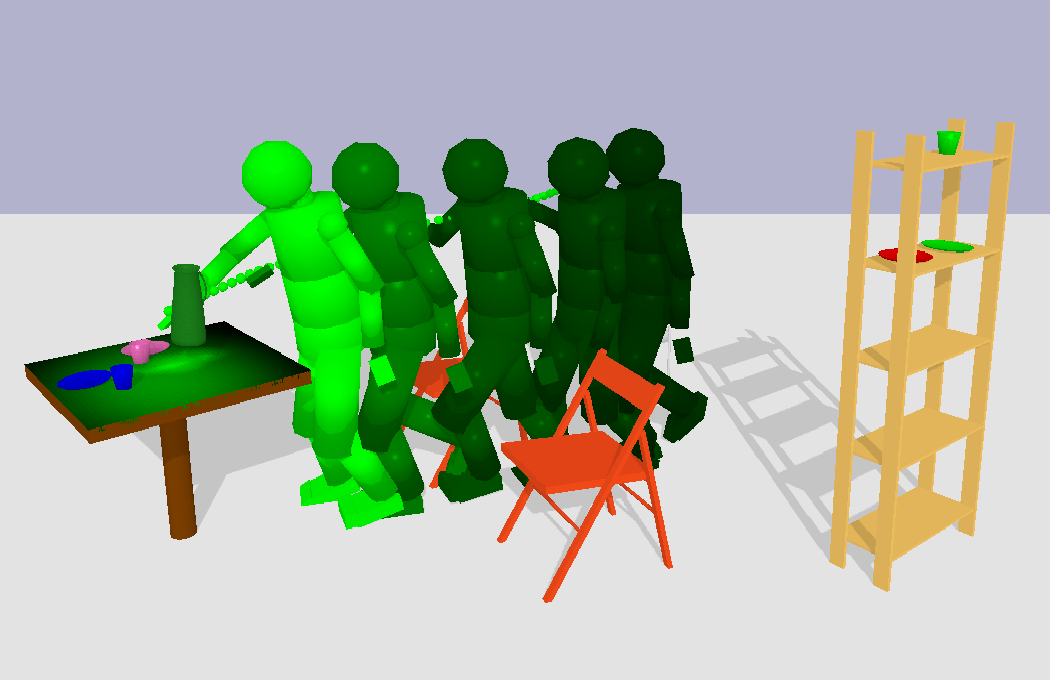}
\includegraphics[width=.32\linewidth]{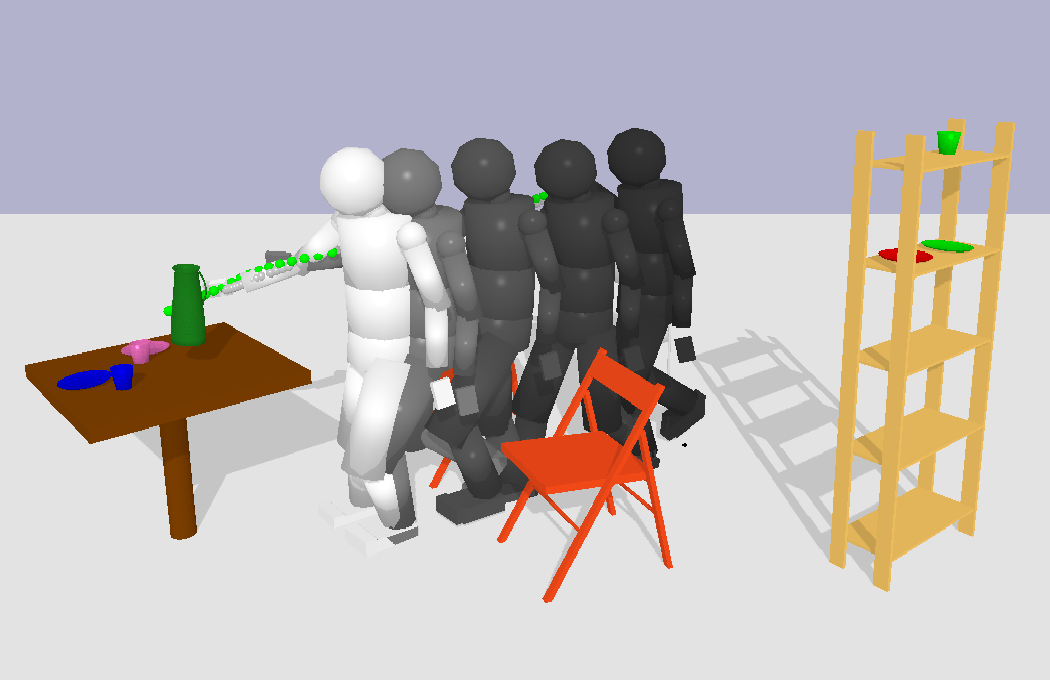}
\captionof{figure}{Prediction for placing a jug on the table. a placement affordance is predicted as a probability density function on the table (a), depicted in green a full-body motion is optimized (b), which is compared to ground truth motion (c).}
\label{fig:place_jug}
\end{center}}]

\thispagestyle{empty}
\pagestyle{empty}

%%%%%%%%%%%%%%%%%%%%%%%%%%%%%%%%%%%%%%%%%%%%%%%%%%%%%%%%%%%%%%%%%%%%%%%%%%%%%%%%
\begin{abstract}

Motion prediction in unstructured environments
is a difficult problem and is essential for safe and efficient
human-robot space sharing and collaboration.
In this work, we focus on manipulation movements in environments
such as homes, workplaces or restaurants, 
where the overall task and environment can be
leveraged to produce accurate motion prediction.
For these cases we propose an algorithmic framework that 
accounts explicitly for the environment geometry based on
a model of affordances and a model of short-term human dynamics 
both trained on motion capture data.
We propose dedicated function networks
for graspability and placebility affordances
and we make use of a dedicated RNN \cite{wang2019vred} 
for short-term motion prediction.
The prediction of grasp and placement probability densities
are used by a constraint-based trajectory optimizer to produce a
full-body motion prediction over the entire horizon.
We show by comparing to ground truth data that
we achieve similar performance for full-body motion predictions
as using oracle grasp and place locations.

\end{abstract}

%%%%%%%%%%%%%%%%%%%%%%%%%%%%%%%%%%%%%%%%%%%%%%%%%%%%%%%%%%%%%%%%%%%%%%%%%%%%%%%%
\section{INTRODUCTION}

When interacting with their environment, humans model the action possibilities directly in the product space of their own capabilities
and the environment.
This idea of the existence of an intuitive and perceptual representation of the possibilities in an environment is known as affordances \cite{gibson1}.

In this paper, we propose an algorithmic framework to learn and encode such affordances from data. By modeling affordances
as probability density functions conditioned
on the environment and the kinematic state of
the human, we are able to anticipate the human intention by
maximum likelihood. This intention can then be combined with
a full-body motion prediction system to produce accurate
predictions as seen in Fig. \ref{fig:place_jug}.

In our experiments grasp and place densities are defined over submanifolds of the hands pose spaces.
Placeability is defined over support planes
and grasps are defined over sphere surfaces
around objects.
Our models for each affordance derive from a common
structure based on recurrent neural networks (RNNs) for modeling the human latent state and dedicated networks, i.e., Convolution Neural Networks (CNNs), for modeling the environment.
The models then combine environment and human latent spaces
using fully connected layers to produce densities
over the sub-manifolds.
Note that we use mixtures to encode multi-modal densities which
is important for placeability.

Given a prediction for placements and grasps
we optimize a full-body movement with
a nonlinear program
\cite{kratzer2019prediction},
which accounts for obstacle and goalset constraints,
and models short-term movements with a data-driven dynamical system.
To the best of our knowledge this paper is the first to
accurately leverage the 3D geometry of the environment
for combined intention and motion prediction
of full-body movements.

We gathered a dataset with 5 participants
using a motion capture system. 
Affordances and short-term motion models were trained
on this dataset. Our results demonstrate superiority of our
affordance densities for predicting placements and grasping
locations. Finally, we show that combining goalset predictions
and motion predictions compares similarly to using
oracle goal locations.

This paper is structured as follows: First we discuss relevant related work in section~\ref{sec:rel_work}. Section~\ref{sec:method} introduces our framework and explains the implementation. Experiments on real motion data are performed in Section~\ref{sec:experiments}. Finally conclusions are drawn in Section~\ref{sec:conclusions}.

\section{RELATED WORK}
\label{sec:rel_work}

\begin{figure*}[t!]
  \centering
  \includegraphics[width=\linewidth]{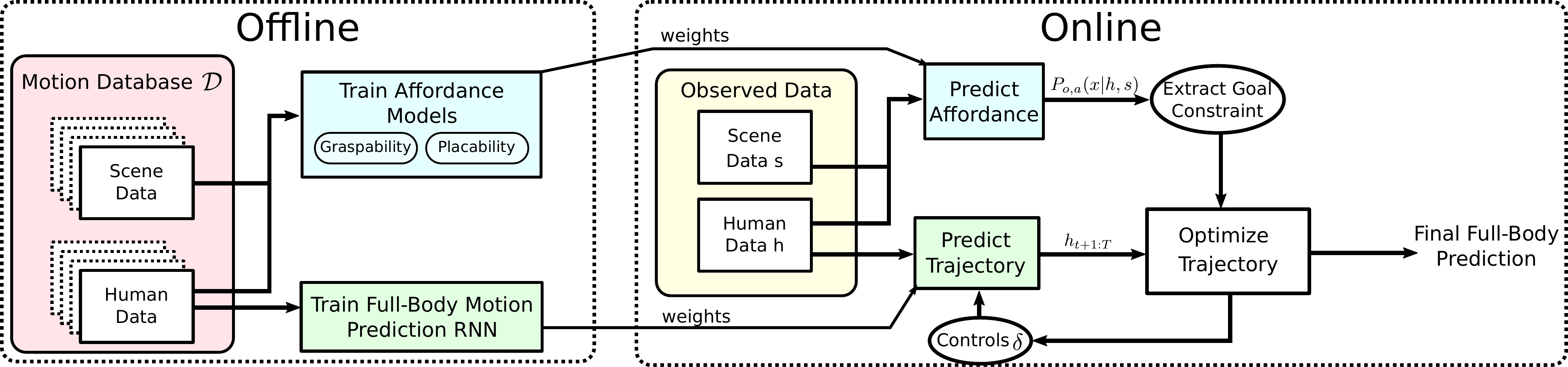}
  \caption{System overview of the proposed method. Offline affordance and full-body motion prediction models are trained. Online a goal constraint is extracted from the affordance and the full-body prediction is optimized to fulfill the constraint.}
  \label{fig:overview}
  \vspace{-.7cm}
\end{figure*}

\subsection{Intention and motion prediction}
In prior work graphical models, such as Hidden Markov Models (HMM) and Conditional Random Fields (CRF), have been used in order to predict human motion or intention. For instance, Bennewitz et al. modeled human intention using HMMs in order to improve navigation behavior of a mobile robot~\cite{bennewitz2005learning}. Kuli{\'c} et al. used HMMs to model full-body motion primitives and applied it to motion imitation~\cite{kulic2012incremental}. Elfring et al. used growing HMMs in order to learn human's goal position from data and use a social forces-based motion model to predict human motion~\cite{elfring2014learning}. Koppula and Saxena focused on movement prediction using conditional random fields~\cite{koppula2016}. While these approaches are sound they generally do not scale to large databases of motion capture or are limited to predict 2d motion of humans and do not deal with the full-body case.

\subsection{Affordances}
The concept of affordances stems its roots from psychology~\cite{gibson1, gibson2} and relates to the action possibilities offered by a
given environment to an animal or human. Jamone et al. present a survey on affordances in the field of psychology, neuroscience and robotics~\cite{jamonesurvey}. The field of visual affordances deals with learning affordances as a computer vision problem~\cite{hassanin2018visual}. Roy et al. use a Convolutional Neural Network based architecture to extract affordance segmentations in RGB images~\cite{multiscale}. Nguyen et al. model affordances using an autoencoder structure~\cite{rgbdcnn}.

In Robotics, affordances can be used to model the actions a robot is able to perform~\cite{montesanoBN, afonso, dAJamone}. For example, Montesano et al. use Bayesian networks to encode affordances and demonstrate how a humanoid robot can use it to interact with objects ~\cite{montesanoBN}.
For Human Robot Interaction affordance models are used to model human action possibilities and to be able to infer human intent~\cite{koppula2013, koppula2016}. Koppula and Saxena define object affordance as potential functions depending upon how the object will be interacted with~\cite{koppula2016}.

In this paper, we design and implement a system to understand human object affordances in a real world table setup task performed in a motion capture environment. As we aim to use the affordance model in order to predict human motion, we use a probabilistic model. Given the human state and the scene context, it predicts a density of interaction possibilities for the corresponding affordance. In particular, we concentrate on graspability and placeability affordances, and model them using a probabilistic neural network framework.

\subsection{Neural Network Human Motion Prediction}
Prior work on full-body human motion prediction for has focused on recurrent neural network (RNN) architectures. Fragkiadaki et al. proposed a Long Short Term Memory (LSTM) based model that is able to train across multiple subjects~\cite{fragkiadaki2015recurrent}.
Martinez et al. introduced a gated recurrent unit (GRU) based approach~\cite{martinez2017human}. A residual connection forces the network to predict velocities and thus improves the generalization capability of the network. Pavllo et al. changed the joint angle representation to quaternions which further improved the predictions~\cite{pavllo2019modeling}. Recently Wang and Feng introduced a position-velocity recurrent encoder-decoder model (VRED)~\cite{wang2019vred}. Their model adds an additional velocity connection as an input to the GRU cell in the recurrent structure.
Motion prediction aproaches based on recurrent neural networks show good results on forecasting of purely human motion. However, they do not handle environmental context, an issue we tackle in this paper due to encoding the environment in our affordance model.

\subsection{Motion Optimization}
Gradient-based optimization algorihtms are widely used in the field of robotics and optimal control~\cite{todorov2005generalized, ratliff2009chomp, toussaint2014newton} for optimizing trajectories.

Moreover, motion optimization techniques have been used for human motion synthesis. For example, Mordatch et al. use motion optimization approaches to synthesize realistic motions and animate human behavior~\cite{mordatch2012discovery, mordatch2013animating}. 

In our prior work we propose to use motion optimization in order to improve short-term motion prediction~\cite{kratzer2018towards, kratzer2019prediction}. We built on the VRED model and used the trajectory optimization technique to change the prediction in order to adapt for specific constraints~\cite{kratzer2019prediction}. In this paper we will use this proposed method in order to predict full-body motion towards a goal state that is sampled from a seperate affordance model. This makes it possible to take environmental context into account.

\section{Combined Intention and Full-body Motion Prediction}
\label{sec:method}

\subsection{Overview}

A schematic overview of the prediction system is shown in Figure~\ref{fig:overview}. Using our captured motion database~$\mathcal{D}$, we offline train probabilistic affordance models  as described in Sections~\ref{ssec:placeability} and \ref{ssec:graspability}. Additionally we train a short-term full-body prediction model. While the affordance models are trained on human data and scene data, the full-body prediction model is only trained on human data. At prediction time, we first use the affordance prediction and extract a goal position. After that a trajectory optimizer is used to iteratively change the predicted trajectory through the full-body model in order to adapt to the goal position, as we describe in Section~\ref{ssec:full-body}. Then the final prediction is returned.

We model affordances by building a relationship between agent $h$ and environment $s$. We aim to find a probabilistic model $P_{o, a}(x|h, s)$ for every object $o$ and action $a$ that gives us a probability over interaction possibilities $x$. For instance, $P_{\text{table, place}}$ would give as probability over possible place locations on the table, while $P_{\text{jug, grasp}}$ would give us a probability over possible wrist locations for grasping a jug.

For the full-body prediction we want to predict a future human trajectory $h_{t+1:T}$ with prediction horizon $T$, based on a previously observed trajectory $h_{0:t}$. We want to constrain it so that the end state $h_T$ fullfills a sample from $P_{o, a}(x|h_t, s)$. For example, the hand of the human should end up at the predicted grasp point or over the predicted place position.

 \begin{figure}[t!]
   \centering
   \begin{subfigure}{\columnwidth}
     \includegraphics[width=\columnwidth]{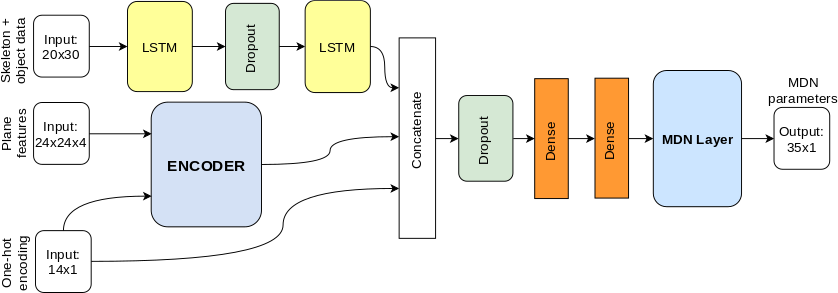}
     \caption{Placeability network architecture}
     \label{fig:place_auto}
   \end{subfigure}
   \begin{subfigure}{\columnwidth}
     \includegraphics[width=\columnwidth]{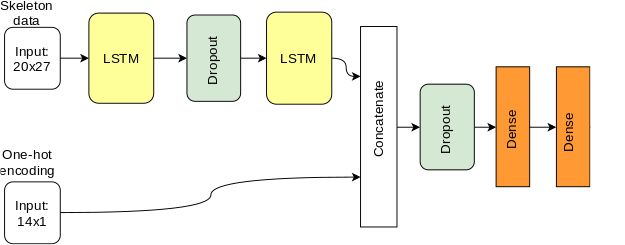}
     \caption{Graspability network architecture}
     \label{fig:base_grasp}
   \end{subfigure}
   \caption{Affordance network architectures. The placeability network has additional inputs for plane occupancy.}
   \vspace{-.7cm}
 \end{figure}

\subsection{Placeability Affordance}
\label{ssec:placeability}
We define the placeability affordance as a probability distribution over possible place locations on a surface. We model the placeability affordance using the neural network architecture shown in Figure~\ref{fig:place_auto}. The inputs $d$ to the model are the human skeleton and object states in positions over a trajectory of 1sec (20 timesteps), a 14 dimensional one-hot encoding of both: the object type the human has in the hand and the surface we compute the affordance for, and a grid that covers the plane state.

\begin{figure}[b]
	\centering
	\includegraphics[scale=0.23]{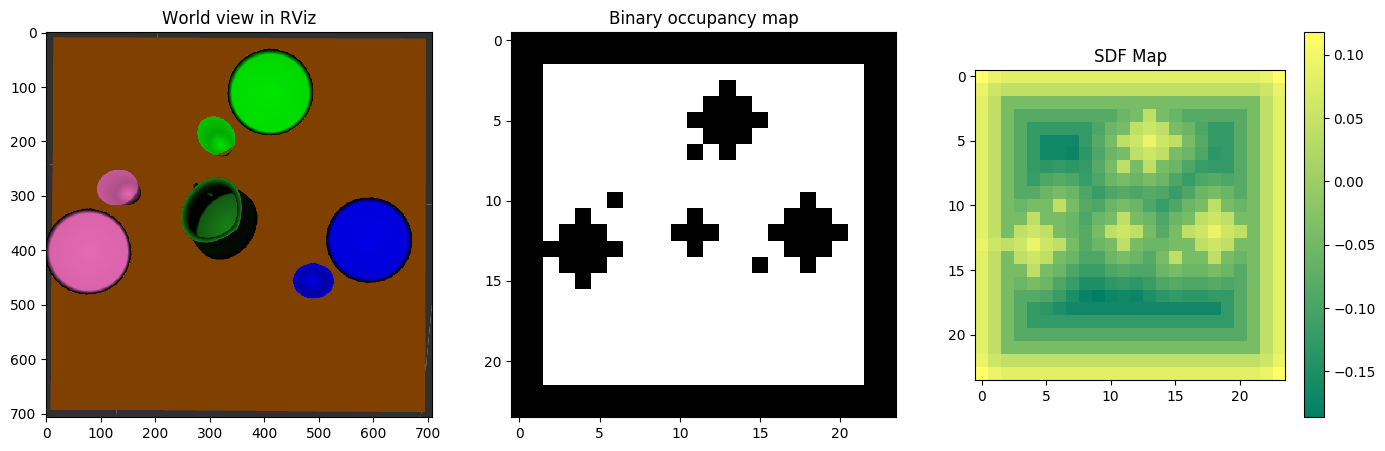}
	\caption{Plane features. From left to right: A visualization of the table with several objects, the corresponding binary occupancy map, a visualization of the signed distance field.}
	\label{fig:place_features}
        \vspace{-.7cm}
\end{figure}

The network additionally takes plane features as input which are
a $24\times24$ grid consisting of a binary occupancy map, a 2d position of the planes reference frame and a signed distance field (SDF) (see Figure~\ref{fig:place_features}).

\paragraph{Multi-modal placements}
Placeability is fundamentally multi-modal.
For instance in our experiments we consider
a table setting scenario such as found in a home or restaurant, four people can sit next to the table, therefore there are four possible locations where the human can place a plate.

A standard approach to model multi-modal distributions
are Mixture Density Networks (MDN)~\cite{mdn}, which we
make use of for modeling placement distributions:
\begin{equation}
\label{eq:mdn_1}
p(x | d) = \sum_{i=1}^m \alpha_i\phi_i(x | d)
\end{equation}
where $m$ indicates the count of the components in the mixture model, $\alpha_i$ are the mixing coefficients. $\phi_i$ are functions representing conditional densities for the $i^{th}$ kernel. 

We use  multivariate Gaussian kernels with diagonal covariance.
We use 7 kernels in output,
which gave good empirical results on our dataset.
The network is trained using a neg-log likelihood (NLL) loss with the 2d place position on the surface as ground truth.

\paragraph{Constraining affordances to free regions}
We improve our placeability model with the intention of making it more robust against violating regions where objects are already placed. We consider 2 approaches to tackle the issue. In the \textit{penalty approach} we modify the cost function to include a penalty term penalizing placement in invalide regions using the value of the SDF map.
In the \textit{transfer learning approach} we learn environment features related to plane occupancy separately.
To achieve this, we build an autoencoder network with inputs being the 4 feature maps and the one-hot encoding vector. The encoder uses two convolutional layers with maxpooling to downsample, the decoder upsamples and uses three convolutional layers. It is trained to output the binary occupancy map of the plane after the object is placed on the plane. We train the autoencoder using a standard mean squared loss.

 The pre-trained encoder model is connected to the main placeability model. The encoder model weights are made non-trainable when the overall placeability network is trained. The intuition here is that, with the autoencoder, we capture the latent representation that are unique for different combinations of the occupancy map. With the pre-trained encoder network producing distinctive feature representations, the main model should learn to not predict outputs in invalid regions.

\subsection{Graspability Affordance}
\label{ssec:graspability}
We model the graspability affordance as follows: \textit{Given that the subject wants to grasp an object of a particular type from its current resting surface, predict the likelihood of the right wrist position for successful grasp action.} The model can then be queried for every object in the scene to get the complete dynamic mapping of the grasp affordance from a human's perspective.

The choice of the posterior distribution influences how the affordance is modeled. We will compare two probability distributions: The Gaussian distribution and the von Mises-Fisher (vMF) distribution. The VMF distribution describes a probability distribution on a hypersphere, which could be useful because grasp points might lie on a hypersphere around the object.

The base structure of our base graspability model is shown in Figure~\ref{fig:base_grasp}. An additional layer is appended in the end, depending on whether we want to model a Gaussian or a vMF distribution.

\paragraph{Gaussian posterior}
In the Gaussian network type, the wrist position is modeled as a 3D position in Euclidean space and the final layer of this network outputs are the parameters of a Gaussian distribution having a diagonal covariance structure.
We use a NLL cost function similar to~\cite{nll}:
\begin{align}
\label{eq:gaussain_cost}
L_{\text{Gauss}}(\theta) &= \frac{1}{N}\sum_{i=1}^N ( \frac{3}{2}\ln(2\pi) + \frac{1}{2}\ln(\text{abs}(C(d_i))) \notag\\
  &+ \frac{1}{2}(y_i - \mu(d_i))^TC^{-1}(d_i)(y_i - \mu(d_i)) )
\end{align}
with $(d_i, y_i)$ being data points and labels and diagonal covariance matrix $C$. The number of output neurons are 6, owing to 3 dimensional mean and variances.

\paragraph{vMF posterior}
The intention for the vMF formulation is to model grasp points as a distribution on a 2D manifold defined on the surface of a sphere:
\begin{align}
\label{eq:vMF}
p_{\text{vMF}}(x; \mu,\kappa)  = C_p(\kappa)\exp(\kappa\mu^Tx)
\end{align}

where $\mu \in  \mathbb{R}^p$, $||\mu|| = 1 $ is the mean direction, with $p=3$ and $\kappa \geq 0$  is the concentration parameter, which defines the spread of the distribution on the surface the hypershere in the direction of $\mu$. $C_p(\kappa)$ is the normalizing constant given by

\begin{align}
\label{eq:norm_vmf}
C_p(\kappa) = \frac{\kappa^{p/2-1}}{(2\pi)^{p/2} I_{p/2-1}(\kappa)^\textrm'}
\end{align}

where $I_s$ denotes the modified Bessel function of the first kind at order $s$.

We have 4 output neurons for the vMF model: 3 for the mean direction and 1 for the $\kappa$~term that defines the spread. Additionally, the output neurons corresponding to the mean direction should satisfy the unit norm constraint. The network is trained on two loss functions simultaneously: neg-log likelihood of the vMF distribution evaluated at ground truth direction and  mean squared loss for the distance parameter.

\subsection{Full-Body Prediction}
\label{ssec:full-body}
The goal for full-body prediction is to find a trajectory of human motion $h_{t+1:T}$ of future states, given a trajectory $h_{0:t}$ of already observed states and our affordance model. For this purpose we use a trajectory prediction framework introduced in our prior work~\cite{kratzer2019prediction}. The framework works in 2 phases: 1) Offline, a VRED model $f$~\cite{wang2019vred} is trained to predict purely kinematic trajectories only based on human motion. $f(h_{0:t}, \delta) = h_{t+1:T}$.  2) Online, trajectory optimization techniques are used to adapt to environmental objectives while being close to the prediction. This is done by changing additional controls~$\delta$ that are added to the VRED architecture.
In this paper we use the  \textit{low-level} objective and the \textit{goalset} objective as described in~\cite{kratzer2019prediction}:
\begin{equation}
    c_{\text{low-level}}(\delta) = \|\delta\|^2
  \end{equation}
The \textit{low-level} objective $c_{\text{low-level}}$ ensures that the deltas are close to zero and therefore the deviation from what the network predicts is small.
\begin{equation}
c_{\text{goalset}}(\delta) = \|\phi_{\text{FK}}(f(h_{0:t}, \delta)_T) - p^*\|^2
\end{equation}
The \textit{goalset} objective $c_{\text{goalset}}$ optimizes the position of the hand of the human to end up close to position $p^*$, with $\phi_{\text{FK}}$ being the forward kinematics map, mapping the last human state to the hand position.

In order to account for our affordance model, we compute the expected prediction position $p^*$ from the affordance model $P_{o, a}(x|h, s)$. Thus, the trajectory will be optimized to end up at this position.

The gradient based optimization algorithm L-BFGS~\cite{byrd1995limited} is used to optimize the trajectory with the loss:
\begin{equation}
  L = \alpha_1c_{\text{low-level}}+\alpha_2c_{\text{goalset}}(\delta)
\end{equation}
with $\alpha$ being hyperparameters. The gradients are calculated using automatic differentiation functionalities from tensorflow.

\begin{table}
 	\centering
 	\begin{tabular}{c c c c c }
 		\toprule

 		& \multicolumn{2}{c}{\textbf{Train set}} & \multicolumn{2}{c}{\textbf{Test set}}                                                                        \\
		\cmidrule(r){2-3}\cmidrule(lr){4-5}

	\textbf{Models}  & \multicolumn{1}{c}{\textbf{NLL}}           & \multicolumn{1}{c}{\textbf{MSE}}          & \multicolumn{1}{c}{\textbf{NLL}}           & \multicolumn{1}{c}{\textbf{MSE}} \\
	\midrule
 		Baseline            & -                                  & 0.0799                                      & -                                 & 0.1051                                                             \\
 		MDN+no CNN features             & -3.1769                                    & 0.0184                                      & -2.6904                                     &  0.0239                                        \\
 		MDN+CNN features                                    & -4.3029                                      & 0.0136                                     & -2.6378                                     & 0.0213                                             \\
   		MDN+CNN+penalty                           & -3.7081                                     & 0.0151                                     & -1.6122                                    & 0.0245                                            \\
  		MDN+transfer learning                             & -4.1168                                      & 0.0115                                     & -2.7793                                     & \textbf{0.0194}                                             \\
  		MDN+transfer+penalty                            & -4.2717                                     & \textbf{0.0107}                                     & -2.4688                                    & 0.0196                                            \\

 		\bottomrule
 	\end{tabular}

 	\caption{Results obtained for Placeability models.}
 	\label{tab:place_table}
        \vspace{-.7cm}
      \end{table}
      
      \begin{figure}
  \centering
  \includegraphics[width=.9\columnwidth]{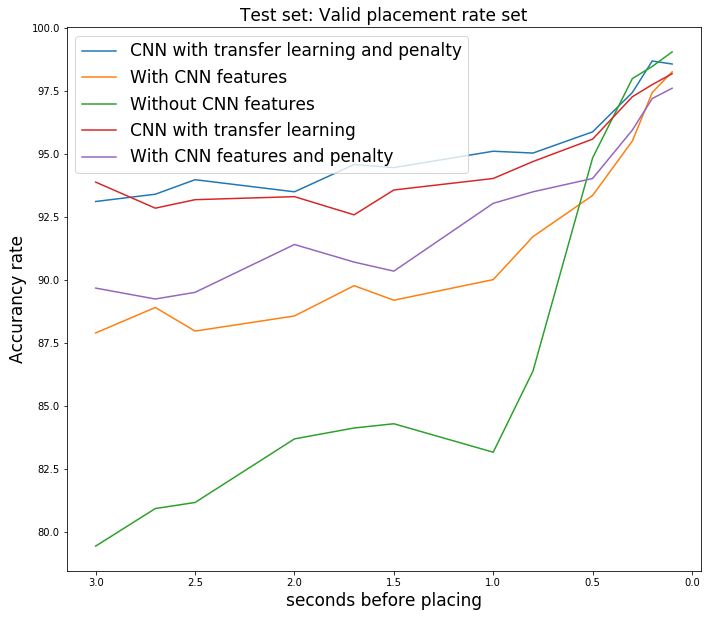}
  \caption{Valid region percentages of placeability over time before the placement happens.}
  \label{fig:place_metric}
  \vspace{-.7cm}
\end{figure}
\section{Results}
\label{sec:experiments}
The models were implemented using Keras~\cite{keras} functional API, with TensorFlow\cite{tf} as backend. To develop the loss functions used to train our custom models, we used the Tensorflow distributions~\cite{tfp} package.
\subsection{Dataset}
In our setup an Optitrack\footnote{https://optitrack.com/} Motion capture system was used. The environment has a total size of $4\times4$ meters. The human subjects were asked to wear a motion capture suit with 50 markers attached to it. There are objects in the scene that are each attached with markers for tracking and can be categorized into two types: The first type of objects are the ones that the users can directly interact with, such as cups, plates, jug and bowl. The second type of objects remain stationary in a given recording session and also acts as supporting bodies over which the first type of objects can be placed, namely a table, a big shelf and a small shelf. We model affordances for the first type of objects. Participants were asked to perform tasks related to setting up the table and clearing it. In the collected data, the users were subject to two affordances, namely graspability and placeability.

A total of 5 users participated in the recording session, with each session being approximately 25 minutes long. We extracted a total of 1551 grasp-place sequences.
For training the models, we split the data based on the subjects. We used data of 3 of the subjects for training and 2 for testing

\subsection{Placeability}

We computed results on different variants of our MDN networks showing the NLL loss and the mean squared error between the mean of the MDN and the ground truth. Results can be seen in Table~\ref{tab:place_table}.  The results are computed on place sequences extracted from the training data. The sequences include place actions for several planes, namely the table and the planes of the big and the small shelf. The baseline for the place affordance is based on a heuristic using the SDF and the distance map. It selects a valid point on the surface which is closest to the human and fits the object. The MDN with transfer learning is our full model. In the MDN+CNN we remove the autoencoder and replace it by two convolutional layers. In the MDN without CNN features no convolutional layers are used at all.
It can be seen that the MDN using the transfer learning technique achieves the best performance on the test set.

 In order to measure whether the model predicts to place into a invalid region (outside of the surface or on space occupied by another object), we additionally calculate the percentage of predictions that satisfy the valid region. 

 Figure~\ref{fig:place_metric} shows the performances of different networks over time before the placement happens.
 Without 2D plane features, the performance of the model is significantly lower compared to the approaches with the CNN network added. It can be seen that the models using the transfer learning or penalty approaches significantly outperform the models without these modifications on the valid placement rates as well as on the Euclidean distance. This holds especially when being farer away from the plane, which is when the prediction of the affordance is most useful.

 The transfer learning approach improves the result by a margin of $5\%$ consistently, along the time axis. This is because, our Autoencoder model inherently learnt unique latent space representation from the CNN features. It was trained to produce a binary occupancy map, that exists in the same space as the input feature maps.

 The pre-trained encoder part produces unique features for the two plane occupancy configurations, thereby it forces the model to predict in the free regions. While the MSE for the model with transfer learning and penalty is about the same as without penalty, training with the penalty term slightly improves the valid region metric for all timesteps.

 \begin{figure}
 	\centering
 	\includegraphics[scale=0.3]{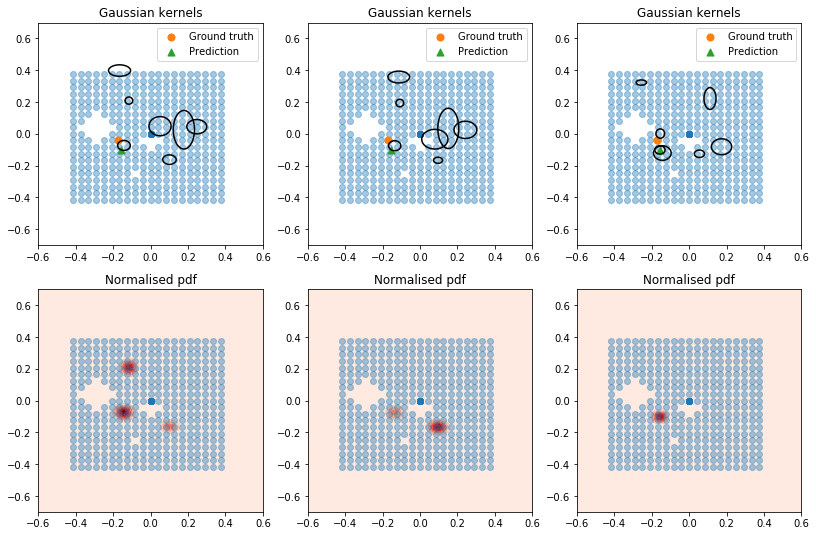}
 	\caption[MDN predictions at different time instances]{MDN predictions at 4s, 1s and 0.5s respectively. The top images show all  7 Gaussian kernels, though most of them have low probability as verified by the density images below.}
 	\label{fig:gmm_pdf_1}
        \vspace{-.7cm}
 \end{figure}

An interesting observation in placeability, is to check the uncertainty in network predictions at different time instances before the object is placed. This can be seen by checking the mixture components predicted by the MDN network and the corresponding density. Figure~\ref{fig:gmm_pdf_1} visualizes the same on a test set example for placing a cup on the table at 3 time instances. It can be seen that when the subject is far away from the table, there are multiple possibilities of potential placeable regions and as the subject moves towards the table, that uncertainty reduces and confines to one dense most likely region.

      \subsection{Graspability}
      We compare the MSE for the vMF model, the Gaussian Model and a baseline. The results can be seen in Table~ \ref{tab:grasp_table}. The baseline for the grasp affordance is based on maximum likelihood, wherein for all combinations of object types and surfaces, the mean distances of the wrist from the object being grasped is computed. During inference, starting from the object, the unit vector along the direction of the right wrist is calculated and the grasp position is computed using this and the corresponding mean distance.

For calculating the MSE with the Gaussian model, the output parameters of the network that correspond to the 3D mean are considered as prediction point, and this is used to compare against the ground truth grasp point. For the vMF model, based on the predicted mean direction and distance, we calculate the 3D position and consider it as the prediction point. Table~\ref{tab:grasp_table} shows the best results obtained from the network. The results are computed at 1s before grasping.

\begin{table}
	\centering
	\begin{tabular}{c c c}
		\toprule
	%	&  \multicolumn{2}{c}{\textbf{Train set}} & \multicolumn{2}{c}{\textbf{Test set}}                                                                        \\
	%	\cmidrule(r){2-3}\cmidrule(lr){4-5}

		\textbf{Models} & \textbf{Train Set MSE} & \textbf{Test Set MSE}\\
		\midrule
		Gaussian model   & 0.0025 & 0.0043 \\
		vMF model  & 0.0042  & 0.0070 \\
		Baseline  & 0.0188  & 0.0250 \\
		\bottomrule
	\end{tabular}

	\caption{Results obtained for Graspability models.}
	\label{tab:grasp_table}
        \vspace{-.7cm}
\end{table}

By observing the MSE, we can see that both neural network models beat the baseline by a significant margin. The Gaussian model achieves a bit better results than the vMF model on the MSE.  However, the models digress in the manner of uncertainty estimation. With the Gaussian model, we get a spherical covariance structure indicating the confidence interval around the mean position. This interval gives the possible locations in 3D space, where the human wrist should position, in order to grasp the object. In case of the vMF model, uncertainty is defined on a 2D manifold, i.e. the surface of a sphere with its center at object centroid, and radius being the predicted distance. The output vMF parameters inform on the direction and spread of the distribution on this manifold. Since this gives a density on a surface around the object, it reflects on the possible approach angle of the wrist for successful grasping.

      \begin{figure*}
  \centering
  \newcommand{\ltscale}{.245}
\newcommand{\vsscale}{.1cm}
\vspace{\vsscale}
\begin{subfigure}{\ltscale\textwidth}
\centering
\includegraphics[width=\linewidth]{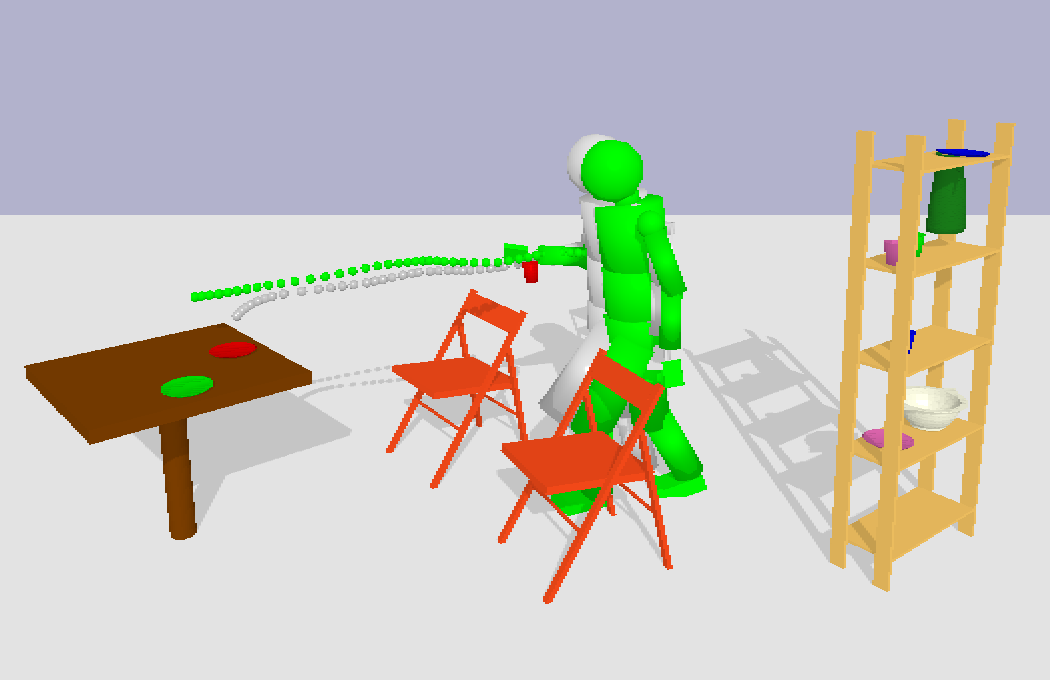}
\end{subfigure}
\begin{subfigure}{\ltscale\textwidth}
\centering
\includegraphics[width=\linewidth]{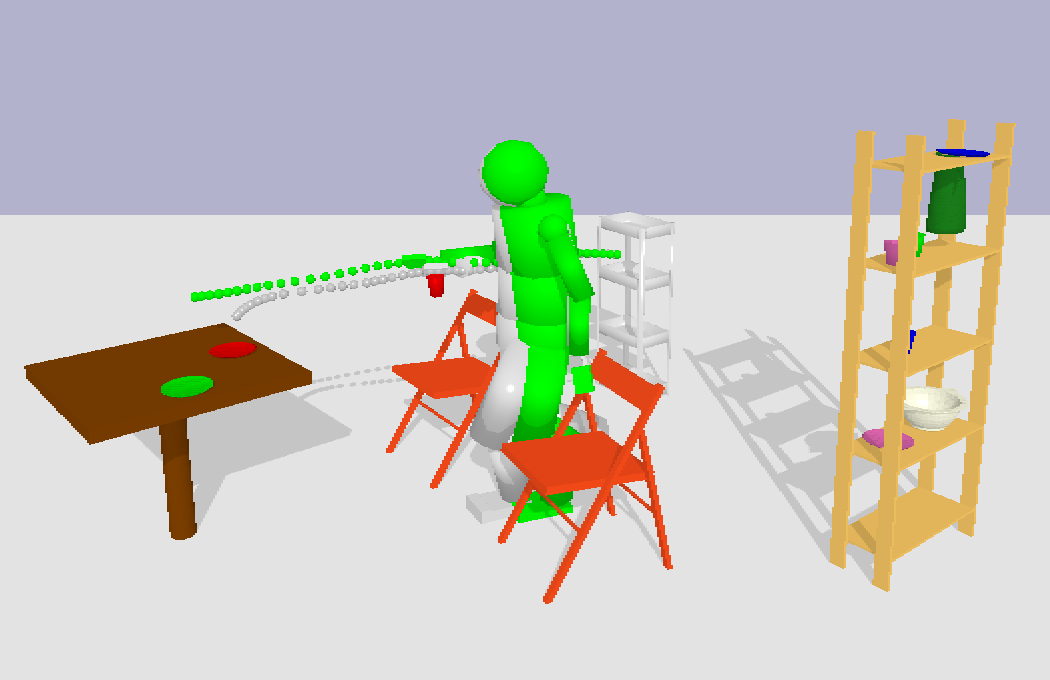}
\end{subfigure}
\begin{subfigure}{\ltscale\textwidth}
\centering
\includegraphics[width=\linewidth]{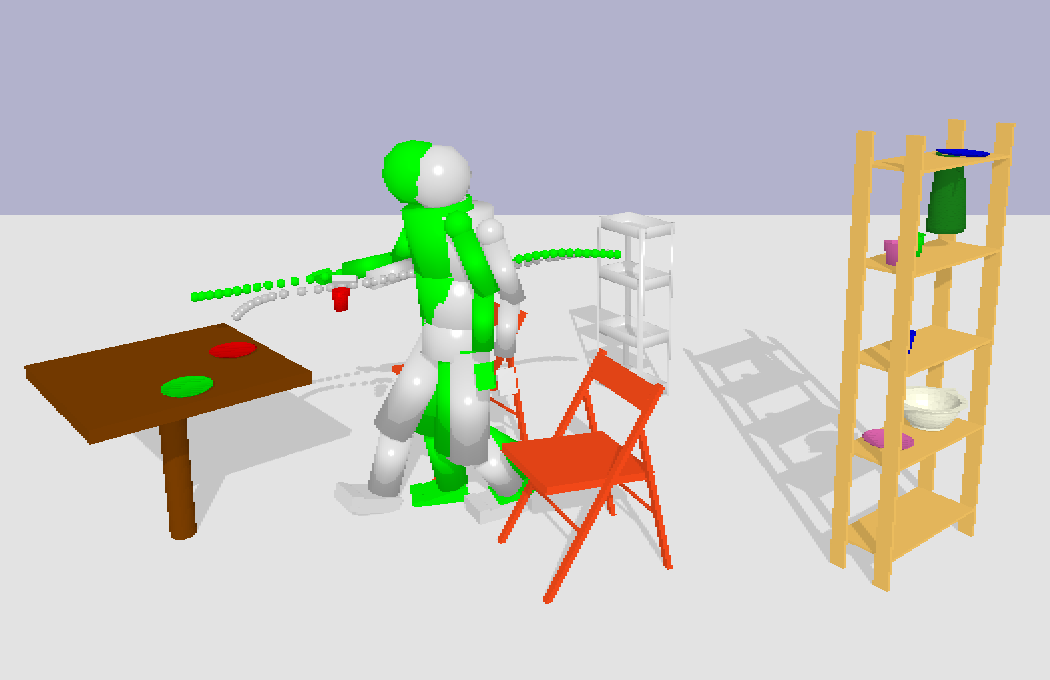}
\end{subfigure}
\begin{subfigure}{\ltscale\textwidth}
\centering
\includegraphics[width=\linewidth]{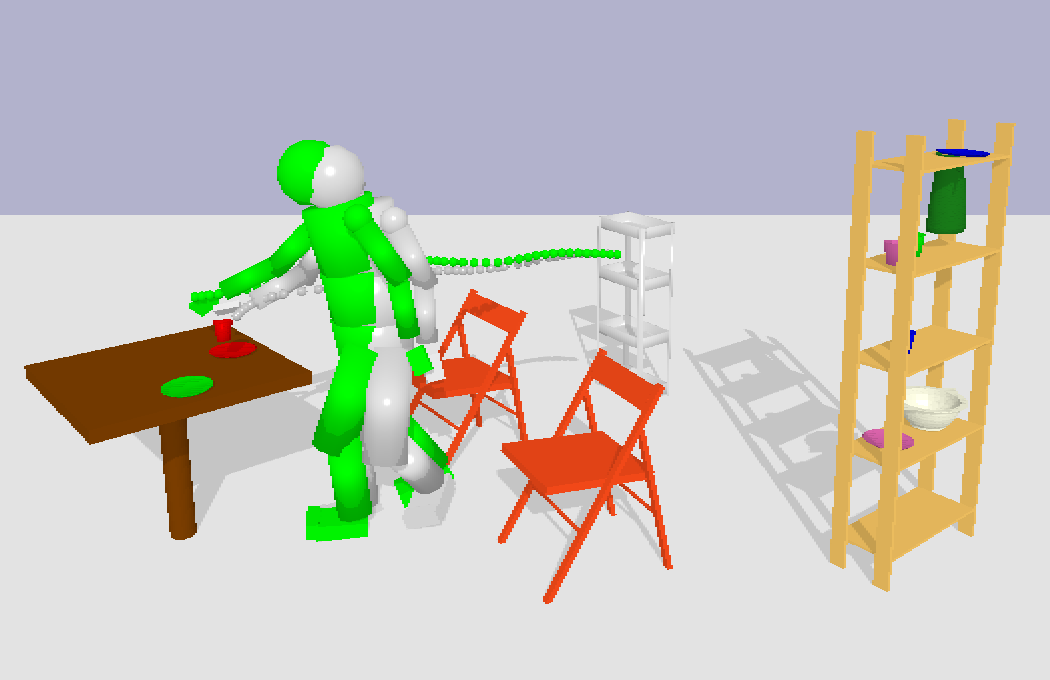}
\end{subfigure}
\\
\vspace{\vsscale}
\begin{subfigure}{\ltscale\textwidth}
\centering
\includegraphics[width=\linewidth]{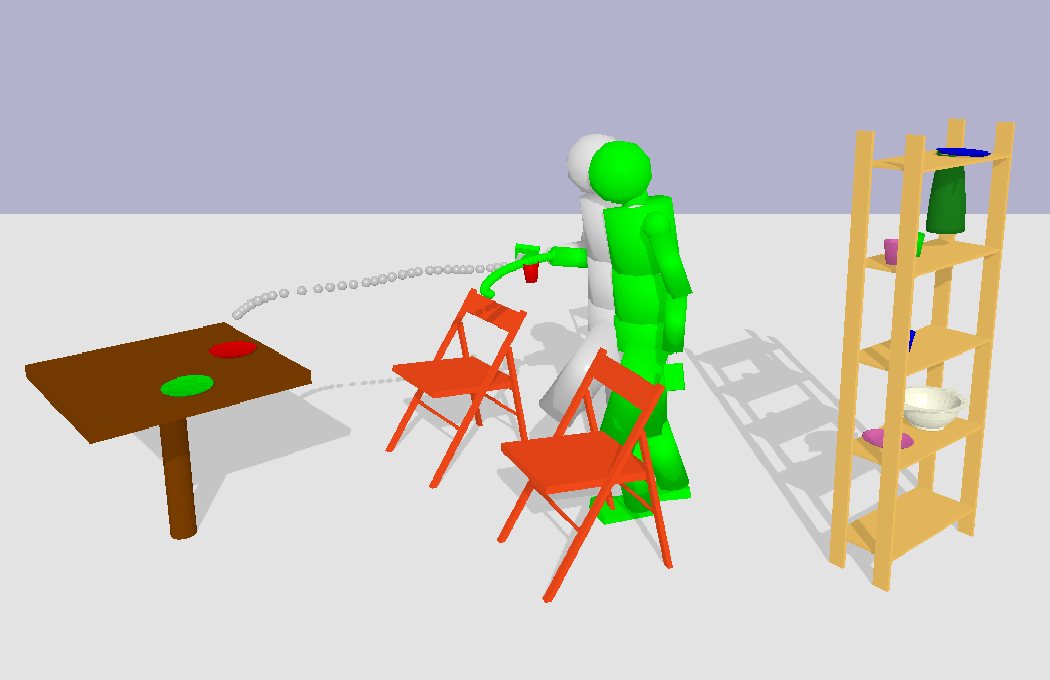}
\end{subfigure}
\begin{subfigure}{\ltscale\textwidth}
\centering
\includegraphics[width=\linewidth]{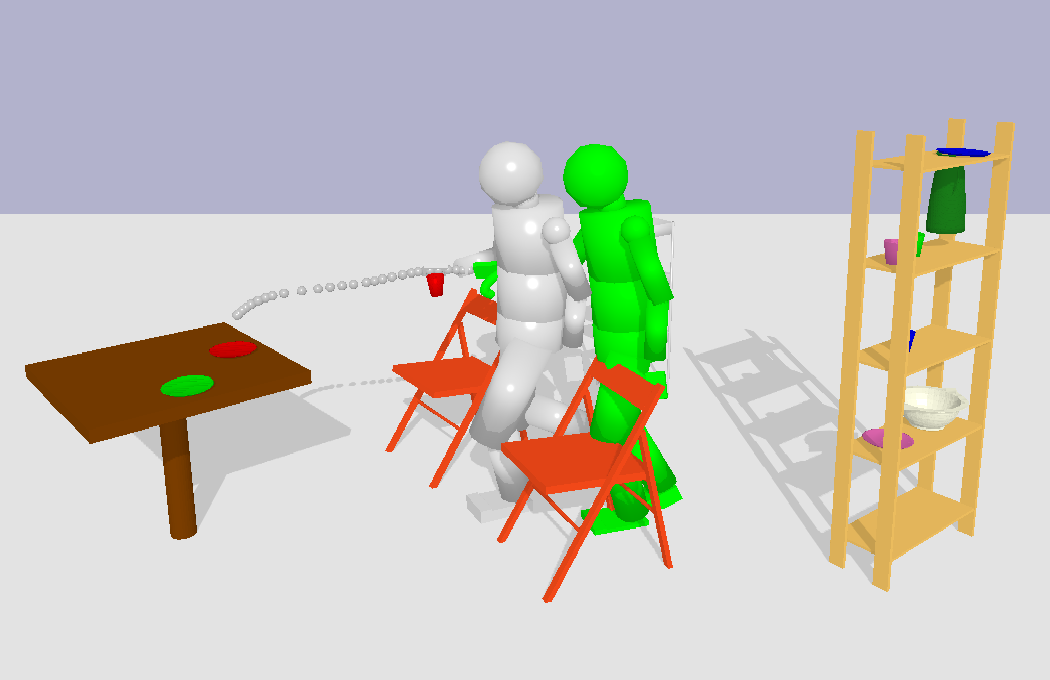}
\end{subfigure}
\begin{subfigure}{\ltscale\textwidth}
\centering
\includegraphics[width=\linewidth]{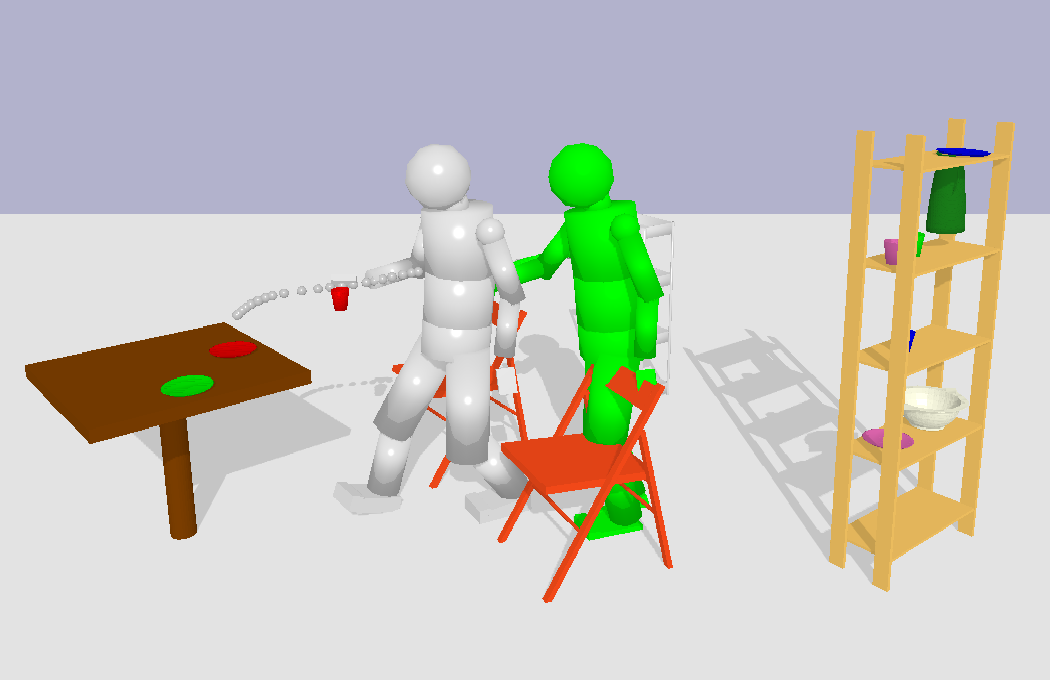}
\end{subfigure}
\begin{subfigure}{\ltscale\textwidth}
\centering
\includegraphics[width=\linewidth]{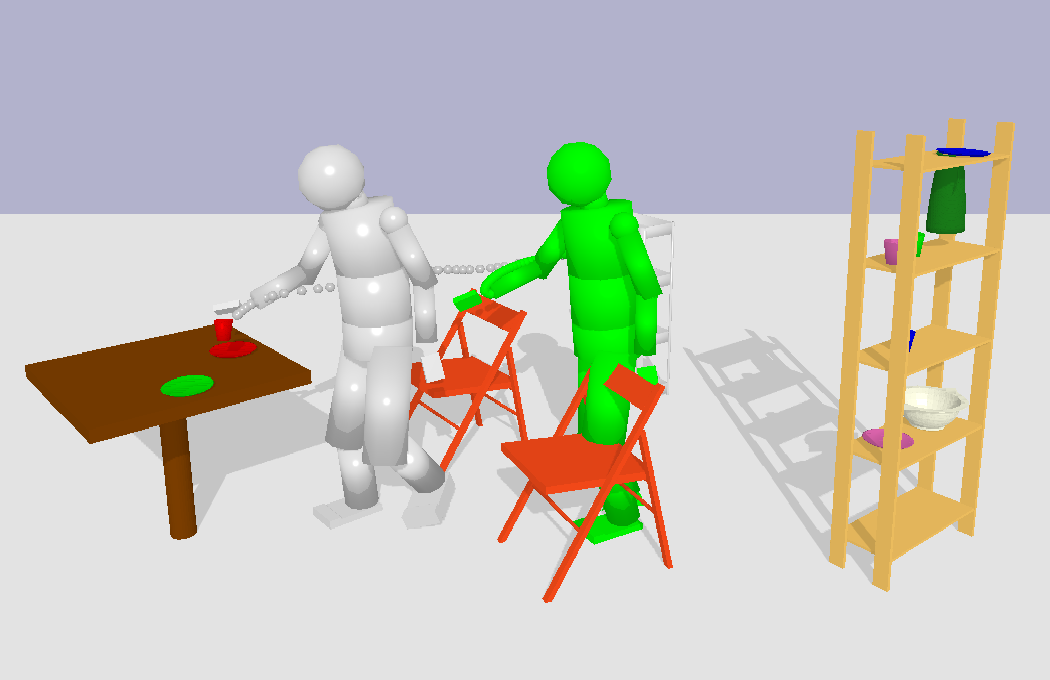}
\end{subfigure}
\caption{Full-body prediction example. From left to right the trajectories after 33ms, 66ms, 100ms and 133ms. Prediction is depicted in green, ground-truth in gray. The top row shows the trajectory with goal optimization towards the affordance, the bottom trajectory is the trajectory without goal optimization.}
\label{fig:example_fb_traj}
\vspace{-.7cm}
\end{figure*}

      \subsection{Full-Body Prediction}
      In order to test the full-body prediction we use the prediction framework introduced in~\cite{kratzer2019prediction}. We train the position-velocity model (VRED) on the training set. From the test data we extract 27 trajectories for placing on the table. We use the place affordance model and extract the expected place point $p^*$ from the MDN. We predict for 1.5sec of motion and optimize the prediction to end up above $p^*$.
      Table~\ref{tab:pred_methods} shows the distance to the ground truth at different times in the future for our method and several baselines. In the first part of the table the sum over distances of key joints (wrists, elbows, knees, ankles and pelvis) is shown, in the second part only the distance of the wrist to the ground truth is shown. Values are averaged over the 27 trajectories.

      The zero velocity baseline just keeps the current state as prediction for future timesteps. The VRED baseline just unrolls the recurrent neural network. Our method takes the affordance prediction into account and optimizes to end up at $p^*$. The oracle has additional oracle information about the true endposition of the wrist.

      It can be seen that the oracle prediction performs best, which is not surprising, as it uses information that is not available at prediction time. Our method using the place point prediction performs second best and outperforms the prediction without any optimization at all time steps.

      Figure~\ref{fig:example_fb_traj} shows an example trajectory for predicting motion to place a cup. The top row shows our method, the bottom row shows a uninformed prediction using VRED. It can be seen that our method is very close to the ground truth, while the uninformed predicts that the human only moves forward a bit and keeps position afterwards.

\begin{table}
  \centering
\setlength\tabcolsep{5.5pt}
  \begin{tabular}{r|c|c|c|c|c|c}
  ms & 250 & 500 & 750 &  1000 & 1250 & 1500\\
    \hline\hline
    Zerovel (b) & 2.38 & 5.18 & 7.76 & 9.68 & 11.09 & 11.94 \\
    VRED (b) & 0.88 & 1.70 & 2.82 & 4.01 & 5.27 & 6.30 \\
    ours (b) &  0.86 & 1.43 & 2.00 & 2.42 & 2.70 & 2.80 \\
    oracle (b) & 0.86 & 1.44 & 1.84 & 2.03 & 2.19 & 2.18 \\
   \hline \hline
    Zerovel (w)  & 0.26 & 0.56 & 0.86 & 1.09 & 1.29 & 1.39 \\
    VRED (w)  & 0.09 & 0.18 & 0.30 & 0.44 & 0.60 & 0.73 \\
    ours (w) & 0.10 & 0.19 & 0.28 & 0.31 & 0.28 & 0.28 \\
    oracle (w) & 0.09 & 0.19 & 0.24 & 0.22 & 0.15 & 0.08 \\
    %Interp (w) &  0.09 & 0.16 & 0.19 & 0.19 & 0.15 & 0.00
  \end{tabular}
  \caption{Error of state prediction per time step for whole body~(b) and right wrist~(w). Reported values are in meters. For the whole body the sum distance of 9 key joints is shown.}
  \label{tab:pred_methods}
  \vspace{-.7cm}
\end{table}

\section{CONCLUSIONS}
\label{sec:conclusions}
We presented a system to learn human object affordances for human motion prediction. We demonstrate that the method can be used to predict full-body trajectories.

A user study was conducted to collect a dataset in a motion-capture setup on a table setup task, in which the actors were subjected to two affordances, namely graspability and placeability.

We modeled the two affordances as conditional probability distributions using deep learning methods by capturing the implicit uncertainty.  For the grasp affordance we use a vMF model. The uncertainty encodes the possible approach angles of the human hand for a successful grasp action. The place affordance was modeled with a MDN model. The uncertainty is encoded as possible regions on the surface where the object can be placed.

Testing within our experimental framework shows the good results of the proposed method.
Furthermore, our experiments proof that the affordances can be used to improve full-body motion prediction within a state-of-the-art motion prediction framework.

\addtolength{\textheight}{-2cm}

\section*{ACKNOWLEDGMENT}
This work was funded by the University of Stuttgart and
the regional research alliance of Baden-W{\"u}rttemberg ``System Mensch''
funded by the German Federal Ministry for Science, Research and Arts.
The authors thank the International Max Planck Research School
for Intelligent Systems (IMPRS-IS) for supporting Philipp Kratzer.

\bibliographystyle{IEEEtran}
\balance
\bibliography{bibliography}

\begin{thebibliography}{10}
\providecommand{\url}[1]{#1}
\csname url@rmstyle\endcsname
\providecommand{\newblock}{\relax}
\providecommand{\bibinfo}[2]{#2}
\providecommand\BIBentrySTDinterwordspacing{\spaceskip=0pt\relax}
\providecommand\BIBentryALTinterwordstretchfactor{4}
\providecommand\BIBentryALTinterwordspacing{\spaceskip=\fontdimen2\font plus
\BIBentryALTinterwordstretchfactor\fontdimen3\font minus
  \fontdimen4\font\relax}
\providecommand\BIBforeignlanguage[2]{{%
\expandafter\ifx\csname l@#1\endcsname\relax
\typeout{** WARNING: IEEEtran.bst: No hyphenation pattern has been}%
\typeout{** loaded for the language `#1'. Using the pattern for}%
\typeout{** the default language instead.}%
\else
\language=\csname l@#1\endcsname
\fi
#2}}

\bibitem{wang2019vred}
H.~Wang and J.~Feng, ``Vred: A position-velocity recurrent encoder-decoder for
  human motion prediction,'' \emph{arXiv preprint arXiv:1906.06514}, 2019.

\bibitem{gibson1}
J.~J. Gibson, ``The senses considered as perceptual systems.'' 1966.

\bibitem{kratzer2019prediction}
P.~Kratzer, M.~Toussaint, and J.~Mainprice, ``Prediction of human full-body
  movements with motion optimization and recurrent neural networks,'' in
  \emph{IEEE Int. Conf. Robotics And Automation (ICRA)}, 2020.

\bibitem{bennewitz2005learning}
M.~Bennewitz, W.~Burgard, G.~Cielniak, and S.~Thrun, ``Learning motion patterns
  of people for compliant robot motion,'' \emph{The Int. Journal of Robotics
  Research}, vol.~24, no.~1, pp. 31--48, 2005.

\bibitem{kulic2012incremental}
D.~Kuli{\'c}, C.~Ott, D.~Lee, J.~Ishikawa, and Y.~Nakamura, ``Incremental
  learning of full body motion primitives and their sequencing through human
  motion observation,'' \emph{The Int. Journal of Robotics Research}, vol.~31,
  no.~3, pp. 330--345, 2012.

\bibitem{elfring2014learning}
J.~Elfring, R.~Van De~Molengraft, and M.~Steinbuch, ``Learning intentions for
  improved human motion prediction,'' \emph{Robotics and Autonm. Systems},
  vol.~62, no.~4, pp. 591--602, 2014.

\bibitem{koppula2016}
H.~S. Koppula and A.~Saxena, ``Anticipating human activities using object
  affordances for reactive robotic response,'' \emph{IEEE Trans. on Pattern
  Analysis and Machine Intell.}, vol.~38, no.~1, pp. 14--29, 2016.

\bibitem{gibson2}
J.~Gibson, \emph{The Ecological Approach to Visual Perception}, ser. Resources
  for ecological psychology.\hskip 1em plus 0.5em minus 0.4em\relax Lawrence
  Erlbaum Associates, 1979.

\bibitem{jamonesurvey}
L.~{Jamone}, E.~{Ugur}, A.~{Cangelosi}, L.~{Fadiga}, A.~{Bernardino},
  J.~{Piater}, and J.~{Santos-Victor}, ``Affordances in psychology,
  neuroscience, and robotics: A survey,'' \emph{IEEE Trans. on Cognitive and
  Developmental Systems}, vol.~10, no.~1, pp. 4--25, 2018.

\bibitem{hassanin2018visual}
M.~Hassanin, S.~Khan, and M.~Tahtali, ``Visual affordance and function
  understanding: A survey,'' 2018.

\bibitem{multiscale}
A.~Roy and S.~Todorovic, ``A multi-scale cnn for affordance segmentation in rgb
  images,'' in \emph{European Conf. on Computer Vision (ECCV)}, vol. 9908,
  2016, pp. 186--201.

\bibitem{rgbdcnn}
A.~{Nguyen}, D.~{Kanoulas}, D.~G. {Caldwell}, and N.~G. {Tsagarakis},
  ``Detecting object affordances with convolutional neural networks,'' in
  \emph{IEEE/RSJ Int. Conf. on Intel. Robots And Systems (IROS)}, 2016, pp.
  2765--2770.

\bibitem{montesanoBN}
L.~{Montesano}, M.~{Lopes}, A.~{Bernardino}, and J.~{Santos-Victor}, ``Learning
  object affordances: From sensory--motor coordination to imitation,''
  \emph{IEEE Trans. Robotics}, vol.~24, no.~1, pp. 15--26, 2008.

\bibitem{afonso}
A.~{Gonçalves}, G.~{Saponaro}, L.~{Jamone}, and A.~{Bernardino}, ``Learning
  visual affordances of objects and tools through autonomous robot
  exploration,'' in \emph{IEEE Int. Conf. on Autonm. Robot Systems and
  Competitions (ICARSC)}, 2014, pp. 128--133.

\bibitem{dAJamone}
A.~Dehban, L.~Jamone, A.~Kampff, and J.~Santos-Victor, ``Denoising
  auto-encoders for learning of objects and tools affordances in continuous
  space,'' in \emph{IEEE Int. Conf. Robotics And Automation (ICRA)}, 2016, pp.
  4866--4871.

\bibitem{koppula2013}
H.~S. Koppula, R.~Gupta, and A.~Saxena, ``Learning human activities and object
  affordances from rgb-d videos,'' \emph{The Int. Journal of Robotics
  Research}, vol.~32, no.~8, pp. 951--970, 2013.

\bibitem{fragkiadaki2015recurrent}
K.~Fragkiadaki, S.~Levine, P.~Felsen, and J.~Malik, ``Recurrent network models
  for human dynamics,'' in \emph{IEEE Int. Conf. on Computer Vision (ICCV)},
  2015, pp. 4346--4354.

\bibitem{martinez2017human}
J.~Martinez, M.~J. Black, and J.~Romero, ``On human motion prediction using
  recurrent neural networks,'' in \emph{IEEE Conf. on Computer Vision and
  Pattern Recognition (CVPR)}.\hskip 1em plus 0.5em minus 0.4em\relax IEEE,
  2017.

\bibitem{pavllo2019modeling}
D.~Pavllo, C.~Feichtenhofer, M.~Auli, and D.~Grangier, ``Modeling human motion
  with quaternion-based neural networks,'' \emph{Int. Journal of Computer
  Vision}, pp. 1--18, 2019.

\bibitem{todorov2005generalized}
E.~Todorov and W.~Li, ``A generalized iterative lqg method for locally-optimal
  feedback control of constrained nonlinear stochastic systems,'' in
  \emph{American Control Conference (ACC)}.\hskip 1em plus 0.5em minus
  0.4em\relax IEEE, 2005, pp. 300--306.

\bibitem{ratliff2009chomp}
N.~Ratliff, M.~Zucker, J.~A. Bagnell, and S.~Srinivasa, ``Chomp: Gradient
  optimization techniques for efficient motion planning,'' in \emph{IEEE Int.
  Conf. Robotics And Automation (ICRA)}.\hskip 1em plus 0.5em minus 0.4em\relax
  IEEE, 2009, pp. 489--494.

\bibitem{toussaint2014newton}
M.~Toussaint, ``Newton methods for k-order markov constrained motion
  problems,'' \emph{arXiv preprint arXiv:1407.0414}, 2014.

\bibitem{mordatch2012discovery}
I.~Mordatch, E.~Todorov, and Z.~Popovi{\'c}, ``Discovery of complex behaviors
  through contact-invariant optimization,'' \emph{ACM Trans. on Graphics},
  vol.~31, no.~4, pp. 1--8, 2012.

\bibitem{mordatch2013animating}
I.~Mordatch, J.~M. Wang, E.~Todorov, and V.~Koltun, ``Animating human lower
  limbs using contact-invariant optimization,'' \emph{ACM Trans. on Graphics},
  vol.~32, no.~6, pp. 1--8, 2013.

\bibitem{kratzer2018towards}
P.~Kratzer, M.~Toussaint, and J.~Mainprice, ``Towards combining motion
  optimization and data driven dynamical models for human motion prediction,''
  in \emph{IEEE-RAS Int. Conf. on Humanoid Robots (Humanoids)}.\hskip 1em plus
  0.5em minus 0.4em\relax IEEE, 2018, pp. 202--208.

\bibitem{mdn}
C.~M. Bishop, ``Mixture density networks,'' 1994.

\bibitem{nll}
D.~A. {Nix} and A.~S. {Weigend}, ``Estimating the mean and variance of the
  target probability distribution,'' in \emph{IEEE Int. Conf. on Neural
  Networks (ICNN)}, vol.~1, 1994, pp. 55--60 vol.1.

\bibitem{byrd1995limited}
R.~H. Byrd, P.~Lu, J.~Nocedal, and C.~Zhu, ``A limited memory algorithm for
  bound constrained optimization,'' \emph{SIAM Journal on Scientific
  Computing}, vol.~16, no.~5, pp. 1190--1208, 1995.

\bibitem{keras}
F.~Chollet \emph{et~al.}, ``Keras,'' \url{https://keras.io}, 2015.

\bibitem{tf}
M.~Abadi \emph{et~al.}, ``Tensorflow: Large-scale machine learning on
  heterogeneous distributed systems,'' 2016.

\bibitem{tfp}
J.~V. Dillon \emph{et~al.}, ``Tensorflow distributions,'' 2017.

\end{thebibliography}

\end{document}